%% file: main.tex
\pdfoutput=1

\documentclass[11pt]{article}
\usepackage[]{ACL2023}

\usepackage{float}
\restylefloat{table}

\definecolor{ForestGreen}{RGB}{34,139,34}
\newcommand{\gu}{\textcolor{ForestGreen}{$\uparrow$}}
\newcommand{\rd}{\textcolor{red}{$\downarrow$}}

\newcommand{\gU}{\textcolor{ForestGreen}{$\Uparrow$}}
\newcommand{\rD}{\textcolor{red}{$\Downarrow$}}

\usepackage[caption=false]{subfig}
\usepackage{enumitem}

\usepackage{diagbox}
\usepackage{graphics}
\usepackage{graphicx}
\usepackage{amsmath}
\usepackage{multirow}
\usepackage[linesnumbered,lined,boxed,commentsnumbered,ruled,vlined]{algorithm2e}
\usepackage[font=small,labelfont=bf]{caption}

\usepackage{booktabs}

\usepackage{times}
\usepackage{latexsym}

\usepackage[T1]{fontenc}

\usepackage[utf8]{inputenc}

\usepackage{microtype}

\usepackage{inconsolata}

\usepackage[english]{babel}
\usepackage{hyperref}

\addto\extrasenglish{

}
\title{In-context Learning as Maintaining Coherency: A Study of On-the-fly Machine Translation Using Large Language Models}

\author{Suzanna Sia \\
  Johns Hopkins University \\
  \texttt{ssia1@jhu.edu} \\\And
  Kevin Duh \\
  Johns Hopkins University \\
  \texttt{kevinduh@cs.jhu.edu} \\}

\begin{document}
\maketitle

\begin{abstract}
The phenomena of in-context learning has typically been thought of as "learning from examples". In this work which focuses on Machine Translation, we present a perspective of in-context learning as the desired generation task maintaining coherency with its context, i.e., the prompt examples. We first investigate randomly sampled prompts across 4 domains, and find that translation performance improves when shown in-domain prompts. Next, we investigate coherency for the in-domain setting, which uses prompt examples from a moving window. We study this with respect to other factors that have previously been identified in the literature such as length, surface similarity and sentence embedding similarity. Our results across 3 models (GPTNeo2.7B, Bloom3B, XGLM2.9B), and three translation directions (\texttt{en}$\rightarrow$\{\texttt{pt, de, fr}\}) suggest that the long-term coherency of the prompts and the test sentence is a good indicator of downstream translation performance. In doing so, we demonstrate the efficacy of In-context Machine Translation for on-the-fly adaptation.  
\end{abstract}

\section{Introduction}
In-context Machine Translation is a relatively new paradigm that uses large autoregressive Language Models to carry out the task of Machine Translation (MT) by being shown translation pairs in the prefix.  From a practitioner's viewpoint, In-context learning presents itself as an attractive approach for rapidly adapting a Translation model on-the-fly. Previous strategies for adapting a pre-trained MT model still require  additional engineering or training of the model, e.g fine-tuning with in-domain data using adaptor layers \cite{philip-etal-2020-monolingual}. Instead, simply changing the inputs to the model might be an effective way to adapt on-the-fly without any model modification. 

The in-context learning paradigm describes a phenomena where large autoregressive language models perform a task when shown examples (known as prompts) in the prefix \cite{brown2020language, bommasani2021opportunities}. Previous work approaches the role of the prompt context as allowing the model to "learn by examples". This intuitive approach to formulating the task of prompt selection has led to the suggestion of selecting examples that are similar to the source sentence being translated. Semantic similarity based on sentence embeddings \cite{liu2021makes} and BM25 have been proposed to select examples to present as ``demonstrations" \cite{rubin2021learning}. This approach was further expanded by \citet{agrawal2022context} who show that BM25 and a heuristic version optimizing for word coverage, is effective for selecting examples.

We focus on Machine Translation as a complex conditional generation task and offer an alternate perspective: \textbf{the in-context paradigm depends on maintaining \textit{coherency}.} Coherence is an aspect of natural language that reflects the overall semantic and syntatic consistency in a body of text \cite{flowerdew2009lexical}. We investigate this by first exploring the model's behavior when showing matching and mismatching domains in the context and the test sentence. Next we consider a stricter notion of coherency using a moving window of previous gold translations directly preceding the test source sentence to be next translated. Our experiments compare the coherence factor with similarity based factors for prompt selection, additionally controlling for length \cite{xie2021explanation} which is typically overlooked but is important to consider for performance and available labeling (translation) budget. The contributions of this work are

\input{tables/format_example}

\begin{itemize}[leftmargin=*]
\item We identify coherency of prompt examples with respect to test sentence as a critical factor for translation performance. Experiments across 3 models (GPTNeo2.7B, Bloom3B, XGLM2.9B) and 4 domains (Medical, Social Media, Wikipedia, and TED Talks) suggest that models perform better when prompts are randomly drawn from the same domain.  
 
\item Within the TED talks domain, we investigate local coherence using document-level translation experiments, by adopting a moving window directly preceding the test source sentence to be translated. Overall, our results across the 3 models and three translation directions (\texttt{en}$\rightarrow$\{\texttt{pt, de, fr}\}) suggest that the coherency of the prompts with regard to the test sentence is a good indicator of translation performance. 

\end{itemize}

\section{Preliminaries}
\subsection{In-context Machine Translation}

In an in-context learning setup, several formatting decisions need to be made on how to present the prompt examples to the model. We adopt the following commonly used prompt format where the instructions are straightforwardly provided as in the following (\autoref{tab:format_example}).\footnote{We also experiment with a different separator "=" used in \cite{lin2021few} (instead of ``English" and ``French"), but find that this does not perform significantly better.} In this work, we consider both sentence level translation (\autoref{sec:exp_domain}) and an on-the-fly document-level setting (\autoref{sec:exp_coherence}).

\subsection{Coherence in Natural Language Text}
\label{sec:prelim_coherence}

The computational linguistics literature holds many competing definitions of coherence in text \cite{wang2014short}. We consider two aspects of coherence, first from a more global level where we investigate domain effects, and also from a local sentence level, where we consider a coherent context as a moving window of previous (gold) translations which directly precede a test sentence. A similar working definition of coherence has been used in discrimination tasks that require a model to identifying the right order of (shuffled) sentences \cite{elsner2007unified, barzilay2008modeling, laban2021can}.

\section{Factors which affect In-context MT}
\label{sec:factors}
We outline several factors studied in this paper related to example selection for In-context MT. 
While we emphasise the notion of \textit{Coherence} (\autoref{sec:prelim_coherence}), by studying the domain factor (\autoref{sec:factor_domain}) and local coherence (\autoref{sec:factor_coherence}), our experiments seek to compare this against other factors that have been highlighted in previous literature. Namely, length (\autoref{sec:factor_length}), surface similarity (\autoref{sec:factor_surfacesim}) and semantic similarity (\autoref{sec:factor_nn}). To demonstrate, in \autoref{tab:format_example}, the first sentence is semantically similar and the second sentence has surface similarity with the test sentence. 

\begin{figure*}[!h]
\centering
\includegraphics[width=1.7\columnwidth]{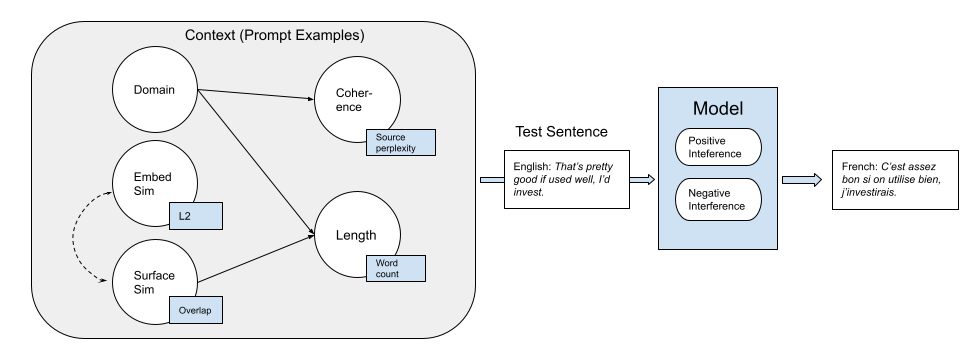}
    \caption{Factors identified and studied in this paper. Selecting from matching Domain increases coherence (\autoref{appendix:domain_ppl}) and each domain has different length distributions (\autoref{sec:domain_vs_length}). Surface similarity and embedding similarity are associated (\autoref{tab:doc_level_results}). Surface similarity selection also results in longer sentences (\autoref{sec:sim_vs_length}) Rectangle boxes next to the node are measures of these factors. We describe and quantify positive and negative interference of the model for translation performance in \autoref{sec:interference}.}
\end{figure*}

\subsection{Length (Translation Budget)}
\label{sec:factor_length}

One previously overlooked factor for in-context MT is the length (number of words) in the prompt examples. The perspective of In-context Learning as implicit Bayesian Inference argues that longer examples provide more evidence to the model on the desired task pattern \cite{xie2021explanation}. Longer examples are also more likely to contain non-trivial translation exemplars, although it is not clear whether this affects downstream performance. We find example length to be correlated with the domain (\autoref{fig:histogram_length}), and it may thus be a confounding factor for in-context MT.

\paragraph{Controlling for Length} 

We adopt the notion of a ``Translation Budget" which is the total word count of all the prompt examples provided (excluding the test sentence). Examples can be selected as long as they satisfy the budget constraint. A generalized algorithm is provided in \autoref{sec:budget_constraint}. From a resource perspective, this reflects the work of the human annotator in providing example translations.

\subsection{Surface Similarity}
\label{sec:factor_surfacesim}
\subsubsection{BM25}
 \label{sec:bm25} 

BM25 \cite{robertson2009probabilistic} is a bag-of-words unsupervised retrieval function that ranks a set of documents based on the query terms appearing in the documents. \citet{agrawal2022context} report that using BM25 to retrieve similar prompt examples outperforms random selection. They also advocate for a variant of BM25 with increased coverage of test sentence source words although with marginal gains (<1 BLEU point) increase. Following \citet{agrawal2022context}, we order the examples according to their similarity to the source, with the most similar examples on the left in all our experiments. 

\subsubsection{Maximising Surface Similarity Coverage} 
\label{sec:bm25-s}
To maximise word overlap across all prompts and the source sentence, we adopt Submodular optimisation by Maximal Marginal Relevance \cite{carbonell1998use,lin2010multi}. Formally we are given a finite size set of objects $U$ (the size of the prompt bank). A valuation function $f:2^U\rightarrow \mathcal{R}_+$ returns a non-negative real value for any subset $X \subset U$. The function $f$ is said to be submodular if it satisfies the property of ``diminishing returns", namely, for all $X \subset Z$ and $Z \notin U$, we have $f(X \cup u) - f(X) \geq f(Z \cup u) - f(Z)$. The algorithm optimises for sentences with maximal word overlap weighted by the BM25 score.  

\subsection{Semantic Similarity (Nearest Neighbors)}
\label{sec:factor_nn}
The semantic similarity of prompts based on their sentence embeddings has also been advocated for selecting good in-context examples. \citet{liu2021makes} apply a pre-trained Roberta-large sentence encoder to the test sentence, and query for its nearest neighbors to use as in-context demonstrations. In our experiments we apply a similar strategy using MPNet base \cite{song2020mpnet} which achieved highest scores on HuggingFace sentence embedding and semantic search benchmarks.\footnote{\url{https://www.sbert.net/docs/pretrained_models.html}} We do not consider training a prompt retriever \cite{rubin2021learning} or fine-tuning the sentence encoder \cite{liu2021makes} in this study, as these are no longer "light-weight" retrieval methods that are comparable with the other unsupervised strategies.

\subsection{Domain Coherence}
\label{sec:factor_domain}
GPT is able to do style transfer just from instructions or from being shown surface prompt examples \cite{reif-etal-2022-recipe}. Simply providing demonstrations from the same domain may induce the large language model (LLM) to generate a similar style which is coherent with the target text. Another possibility is that particular lexical translation exemplars which match the source sentence may be present. However, due to the very high dimensionality of the raw vocabulary, this is less likely if translation examples are randomly sampled. 

Domain may also present spurious correlations which are confounded by the training data of LLMs. For instance, there may be certain domains which are better at eliciting Translation behavior from the model, regardless of what the test domain is. 

\subsection{Local Coherence (Moving Window)}
\label{sec:factor_coherence}
We hypothesise that the local coherence (\autoref{sec:prelim_coherence}) of the context to the test sentence to be translated may be an important factor for performance. To test this, we adopt a moving context window of the previously translated gold sentence pairs as the prompt examples.  To our knowledge, \autoref{sec:factor_domain} and \autoref{sec:factor_coherence} are previously unexplored for In-context Machine Translation.

\section{Experiments}

\subsection{Data}
\label{sec:data}
\paragraph{Domain Coherence}
We organise our experiments investigating four \texttt{en}$\rightarrow$\texttt{fr} 
 domains, WMT19 Biomedical (MED) \citep{bawden-etal-2019-findings}, a social media dataset, MTNT \citep{michel2018mtnt}, multilingual TED Talks, and Wikipedia-based FLORES \cite{flores}. Except for MED, all other datasets have a wide range of topics in the train (prompt bank) and test set which are shuffled in random sampling, and thus the domain experiments are more focused on the writing style of the text.

 We use standard train-test splits, with the trainset being used as the prompt bank. Scores are reported using SacreBLEU \citep{post-2018-call}.\footnote{nrefs:1 $\mid$ case:lower $\mid$ eff:no $\mid$ tok:13a $\mid$ smooth:exp $\mid$ version:2.0.0}

\paragraph{Local Coherence (document level)} 
We use the Multitarget TED Talks dataset from \citet{duh18multitarget}. The original dataset has 30 documents in the test set, where each document corresponds to a 10-20 minute TED talk. To increase the size of the test set, we partition the "original" trainset into a train (prompt bank) and test split, where talks with a minimum of 100 lines were used as the test and talks with less than 100 lines were used as the "out-of-document" prompt bank. We used 120 test documents that had a minimum of 100 lines, and we evaluated each up to 120 lines, where each TED talk is a  document. The document level BLEU scores are reported for three language directions \texttt{en}$\rightarrow$\{\texttt{fr}, \texttt{pt}, \texttt{de}\}. We do not use a dev set as there is no training or any tuning of any hyperparameters. Since this is a non-standardised data split, we provide the numbers in the following table.  

\begin{table}[h!]
    \centering
\resizebox{\columnwidth}{!}{
\begin{tabular}{p{0.30\columnwidth} p{0.23\columnwidth} p{0.23\columnwidth} p{0.15\columnwidth}}

& Talks (Docs) & Lines per doc & Total Lines \\
\midrule
"Outside-doc" Prompt Bank & 450 & <100 & 26000+ \\
\midrule
"Within-doc" Prompt Bank & 1 &  100-120 & 120 \\ 
\midrule
Test  & 120  & 100-120  & 12000+ \\

\end{tabular}}
\end{table}

\subsection{Models}
\label{sec:models}

We use three models, GPTNeo2.7B \citep{gpt-neo}, XGLM2.9B \cite{lin2021few}, and Bloom3B \cite{scao2022bloom} which are open access LLMs available on HuggingFace \cite{wolf-etal-2020-transformers}. The later two have been advertised as "Multilingual Language Models". 
GPTNeo2.7B is a GPT3 replicate pretrained on The Pile \cite{gao2020pile}, while XGLM adopts a similar architecture trained on a multilingual corpus (CC100-XL). Bloom3B has been trained on the ROOTS Corpus \cite{laurenccon2022bigscience}, a collection of huggingface datasets of 1.6 TB of text. To our knowledge, there has not been any reports of sentence level parallel corpora in the training datasets of these models. 

\input{tables/domain_results.tex}

\subsection{Algorithm for Greedy selection with Length Constraint}
\label{sec:budget_constraint}

In our experiments, we investigate \texttt{BM25} (\autoref{sec:bm25}), \texttt{BM25} with submodular optimisation (\texttt{BM25-s}; \autoref{sec:bm25-s}), and semantic similarity (\texttt{nn}; {\autoref{sec:factor_nn}). To control for length effects, we employ an algorithm for selection with length constraints (\autoref{algo1}) which closely follows greedy submodular algorithms \cite{krause2008beyond}. Retrieval methods adopts a utility function: $f$, which is used to retrieve highest scoring sentences. For \texttt{BM25} and \texttt{BM25-s}, $f$is \texttt{BM25}, while $u_i$ is selected by $f(\{u\})$, and $f(\{u\}|X_i)$ respectively. While for \texttt{nn}, $f$ is the L2 embedding similarity between prompt sentence and test query. 

\begin{algorithm}
 \textbf{Input:} (Submodular) function $f:2^U \rightarrow R_{+}$, cost function $m$, budget $b$, finite prompt bank $U$ \\
 \textbf{Output:} $X_k$ where $k$ is the number of iterations/prompts.\\
 Set $X_0 \leftarrow \empty$; $i \leftarrow 0$;\\
 \While{$m(X_i)<b$}{
    $u_i = \textrm{argmax}_{u\in U \backslash X_i} f(\{u\} \mid X_i)$\\
    $X_{i+1} \leftarrow X_i \cup {u_i}$;\\ 
    $i \leftarrow i+1$
 }
 \caption{Generalised greedy (submodular) algorithm with length budget}
 \label{algo1}
\end{algorithm}

\section{Analysis of Factors}
\label{sec:experiments}

\subsection{Domain Coherence [\autoref{tab:domain_results}]}  
\label{sec:exp_domain}

\textit{Does coherence of domain allow models to adapt on the fly?} 
If models are adapting to the domain shown in the context, sampling and testing within the same domain should result in the highest translation performance, as compared to being shown examples out of domain. For example, if we are testing on the TED domain, is it important that the prompt be also drawn from TED or is it sufficient to have sentence pairs from any domain illustrating the translation task? To account for prompt selection and ordering effects, all inference runs were repeated with 5 randomly sampled prompt sets from the training data. We focus on \texttt{en} $\rightarrow$ \texttt{fr} which is common across datasets.

\paragraph{Results and Discussion} 

\begin{itemize}[leftmargin=*]
\item Models are able to perform some form of domain adaptation on-the-fly. There appears to be evidence of domain adaptation in Bloom3B and XGLM, as sampling and testing within the same domain (e.g., sample from MED test with MED) mostly results in the highest performance column-wise. We also observe that matching domains result in lower conditional source sentence perplexity (\autoref{appendix:domain_ppl}).

\item For GPTNeo, sampling from FLORES results in the best translation performance across all test sentences even with domain mismatch. This suggests that translation performance in GPTNeo is best induced using FLORES and is less adaptive to the domain. Note that the second best column wise result for GPTNeo tends to occur when there is matching prompt and test domain.

\end{itemize}

\subsection{Domain controlling for Length}
\label{sec:domain_vs_length}
\textit{How does length of prompts affect translation across different domains?} In \autoref{fig:histogram_length}, we randomly sample 1000 sentences from each domain's training set. Randomly sampled sentences from different domains show distinct length effects. We study the impact of these length effects by selecting either a 5-10 word or 15-20 word long sentences for translation examples, and compare the differences in scores for the non-filtered scenario (\autoref{tab:domain_length}).  

\begin{figure}[!h]
    \centering
    \includegraphics[width=0.8\columnwidth]{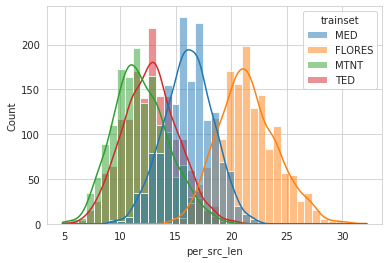}
    \caption{Histograms of sentence lengths (word counts) randomly sampled from different domains, which has implications for the total prompt length when sampling from these domains. FLORES sentences tend to be nearly twice as long as MTNT and TED sentences.}
    \label{fig:histogram_length}
\end{figure}

\input{tables/domain_length.tex}

\paragraph{Results and Discussion}

\begin{itemize}[leftmargin=*]
    \item When source prompt sentences are 5-10 words, all BLEU scores decrease. For 15-20 words sentences which is "long" for MTNT and TED, but "short" for FLORES, the BLEU score of the former increases while the latter decreases. BLEU scores are similar for MED as 15-20 words is close to the mean of MED length distribution.
    
    \item We inspect the length of generation under different prompt lengths, and find that average differences in generation length are marginal (only 1-2 words difference) indicating that poorer performance is not simply due to a difference in generation lengths.
\end{itemize}

\input{tables/doc_level_results.tex}

\subsection{Local Coherence [\autoref{tab:doc_level_results}]}
\label{sec:exp_coherence}
\textit{How important is a coherent context (as compared to other prompt selection methods?)}  \autoref{sec:exp_domain} showed that models are able to adapt when shown prompts from a matching domain. We hypothesise that coherence of the prompts with respect to the test source sentence (\autoref{sec:prelim_coherence}) is an important factor for performance. 

We use the TED talks dataset (data preparation described in \autoref{sec:data}), and consider a moving window of previous gold translations (\texttt{window}) as a coherent context for the model.\footnote{Preliminary experiments using model generated instead of gold translations performed worse than random.} We compare this against the baselines of (\texttt{BM25}; \autoref{sec:bm25}), (\texttt{BM25-s}; \autoref{sec:bm25-s}), and Nearest Neighbor retrieval of sentence embeddings (\texttt{nn}; \autoref{sec:factor_nn}) from a large prompt bank outside the document. We use a prompt set of 5 examples for all experiments, and randomly sample from outside of the document if the available window is smaller than 5. Document level BLEU scores are averaged across 120 documents and reported in \autoref{tab:doc_level_results}.  

\paragraph{Quantifying Similarity}
We report the ROUGE1-precision (\texttt{coverage}; \citet{lin2004rouge}) and the L2 Euclidean distance (\texttt{L2}) of the source sentences in the prompt set, with the test source sentence to be translated. If translation performance is due to word overlap or embedding similarity, then we expect that having a higher \texttt{coverage} or lower \texttt{L2} would have better performance than \texttt{window}. Note that all similarity based retrieval methods depend only on the source sentences, and is model and target language independent. i.e., the single \texttt{coverage} and \texttt{L2} value applies for all results columns in \autoref{tab:doc_level_results}.

\paragraph{Results and Discussion}

\begin{itemize}[leftmargin=*]
\item The moving window (\texttt{window}) outperforms all other baselines across the 3 models and 3 language directions, with the exception of Bloom3B on \texttt{en}$\rightarrow$\texttt{de} direction. The gains are from 0.5 to 2.6 BLEU points from the next best performing retrieval method. Importantly, \texttt{coverage} and \texttt{L2} shows that the performance is not due to similarity or word overlap.  

\item Interestingly, randomly sampling sentences from within the document (talk) performs well compared to other similarity based retrieval methods from outside of the document. This further highlights that coherence is a critical factor for In-context Machine Translation. 

\item Similarity based retrieval mostly does better than randomly sampled prompt sets, which is consistent with existing literature which did not consider the factor of coherence. A notable exception is XGLM \texttt{en}$\rightarrow$\texttt{fr} results, where similarity based methods are doing poorly compared to that reported by \cite{agrawal2022context}. We find that the similarity based retrieval methods does better for XGLM when the number of prompts is increased from 5 to 15. The same trend is observed at 15 prompts, \texttt{window} continues to outperform the other methods (results in \autoref{appendix:exp_coherence}). 
\end{itemize}

Crucially, this set of experiments show that \textit{similarity based methods are not as critical for translation as compared to coherency}, a new factor that we identify in this work.

\subsection{Similarity based Retrieval within the Document}
\label{sec:sim_vs_length}
\textit{How well do similarity based retrieval methods perform for previous on-the-fly translations?}
In \autoref{sec:exp_coherence}, we established that using a moving window (local coherence) outperforms retrieval from outside the document with similarity-based retrieval methods. Here we apply \texttt{bm25}, \texttt{bm25-s}, \texttt{nn} for retrieval \textit{within} the document. We consider the more realistic "on-the-fly" or computer-aided translation scenario, where the human translator works with MT systems, and translation examples in the document can only be selected prior to the test sentence \cite{alabau2014casmacat}.

\paragraph{Controlling for Length}
When doing retrieval based methods within the document for an "on-the-fly" setting, length factors in and longer sentences are retrieved on average. We thus investigate budgeting for the length constraint to be same as the moving window (\texttt{window}). For every test sentence, we compute the budget used by it's own moving window, and apply it as a length constraint to for the other retrieval based methods 
as described in \autoref{sec:budget_constraint}. Results are presented in \autoref{fig:barchart_lengthcontrol}.

\begin{figure}[t!]
\centering
\includegraphics[width=0.8\columnwidth]{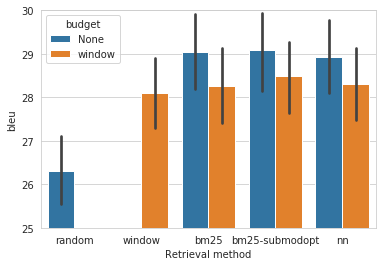}
    \caption{Comparison of Retrieval methods controling for budget: No budget or same budget as moving window. Model is GPTNeo2.7B on en->fr. \texttt{random} is sampled within the document.}
\label{fig:barchart_lengthcontrol}
\end{figure}

\paragraph{Results and Discussion}
\begin{itemize}[leftmargin=*]

\item We observe similar performance for all retrieval methods, with \texttt{bm25-s} doing slightly better than \texttt{bm25} and nearest neighbors (\texttt{nn}).  

\item Without any budget restriction, performance of retrieval methods outperforms \texttt{window}. However when restricted to the same budget as \texttt{window}, we find that the performance is within 0.1-0.5 BLEU score difference. Furthermore, the \texttt{coverage} is only 0.01-0.03 less if not using similarity based retrieval, indicating that most of the differences in contributions could be coming from the length effect and not because of similarity.

\end{itemize} 

\section{Further Analysis and Discussion}
In this section, we focus on GPTNeo2.7B and in the \texttt{en}$\rightarrow$\texttt{fr} direction. 

\subsection{Perplexity and Coverage}
One natural question that arises is the relationship between \texttt{Coverage}, Coherence, and translation performance. Although there is no widely accepted measure of \textit{general coherence}, we can formulate this with respect to the particular model being studied. We consider the model's conditional perplexity of the test sentence given the context. Perplexity is a widely used measure of suprisal in text and has also been used as a measure in topic coherence \cite{newman2010automatic}.  Concurrent work by \citet{gonen2022demystifying} argue that total perplexity of the input sequence is related to In-context performance. 

In \autoref{fig:scatter_plots}, we produce scatterplots of Sentence BLEU scores, source perplexity and \texttt{Coverage} (word overlap). We observe that there is a negative relationship between source perplexity and Sentence BLEU (-0.22 Pearson's r), but very noisy relationship between Sentence BLEU and word overlap, and word overlap and source perplexity.

\begin{figure*}[!h]
\centering
\includegraphics[width=1.7\columnwidth]{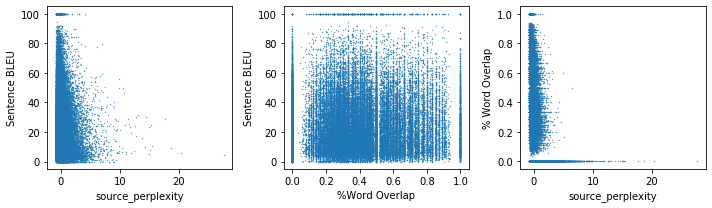}
    \caption{Scatterplots of Sentence BLEU Scores, with Source Perplexity and Word Overlap}
\label{fig:scatter_plots}
\end{figure*}

\subsection{Studying Local Coherence [\autoref{tab:shuffle_doc_level}]}
\label{sec:exp_shuffle}
 We compare the \texttt{window} with other baselines which may give some indication of what is important in the document in terms of local coherence. 

\begin{itemize}[leftmargin=*] 

    \item \texttt{Shuffle} simulates whether the model is affected by the the local coherence by shuffling sentences within \texttt{window}. 
    \item \texttt{Static} refers to the first $k$ (window size) translation sentences of the document which is then held fix throughout when translating the rest of the document. 

\end{itemize} 
\input{tables/shuffle_doc_level}

Interestingly, shuffling the set of prompts within the moving window which breaks the natural ordering of the document "coherence" does not deteriorate in-context translation performance. This finding is consistent across several models and languages \autoref{appendix:shuffle}. The ordering of the document does affect source perplexity, with perplexity increasing from 11.1 $\rightarrow$ to 12.0, however this does not negatively affect translation performance. This suggests that the relationship between coherence and translation is indirect or non-linear, and the way models use context might be counter-intuitive; a view increasingly advocated by recent research \cite{ webson2021prompt, min2022rethinking}. Overall this suggests we may benefit from methods which perform selection from within the document which we leave to future work.

\subsection{Do we need Translation examples at all?}
\label{sec:interference}
Given the rise of instruction-following GPT \cite{ouyang2022training} a reasonable question is whether prompt example selection will still be relevant in future models. For a large language model, merely providing the instruction "Translate English to French" without any prompt examples (zero-shot) can still elicit a translation. In spite of zero-shot success, a common finding (for MT as well as other NLP tasks) is that providing more prompt examples typically results in better performance albeit with diminishing effect. Since examples are not strictly \textit{necessary} for translation but can \textit{enhance} the model's downstream translation ability, what is the role of prompt examples? 

\input{tables/interference.tex}

\paragraph{Positive vs Negative Task Interference}

One curiosity that we observe across all of our experiments, is that prompt sets \textit{do better on-average rather than across all examples}, relative to the Zero-shot, instructions only setting. This suggests a notion of interference;  examples may guide generation towards a poorer translation (negative interference) or better translation (positive interference). A closely related concept is task location \cite{reynolds2021prompt}. 

\autoref{tab:interference} quantifies this across different methods corresponding to the results in \autoref{tab:doc_level_results} for GPTNeo2.7B en$\rightarrow$ fr direction. \texttt{window} has both the highest positive interference and lowest negative interference. From \autoref{fig:histogram_bleu}, the major role of prompting methods compared to the zero-shot scenario is to have greater positive interference chiefly over sentence BLEU of 20-60 and for some extreme cases of 100 BLEU, although a large proportion of sentences still lie in the low-scoring region. 

\begin{figure}
  \centering
\includegraphics[width=0.9\columnwidth]{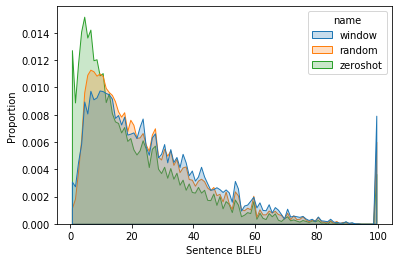}
 \caption{Histograms of sentence BLEU scores for \texttt{zeroshot} (no prompts), \texttt{random}, and \texttt{window}. }
 \label{fig:histogram_bleu}
\end{figure}

\section{Conclusion}

In-context Learning has typically been thought of as learning from examples. In this work, we introduce a different perspective of coherency of the context with the test sentence. We first showed that models are mostly able to adapt to different writing styles when the prompt bank and test set are matching/consistent in domain. Experiments across 3 models and 3 languages show that a moving window is up to 2.6 BLEU points better than previously reported similarity based retrieval methods from outside the document. From this perspective, the problem of prompt selection for in-context MT is one of maintaining a coherency for text generation. Preliminary analysis on local coherence effects, and the presence of negative interference compared to the zero-shot setting, suggests avenues for future work on investigating more careful mechanisms for controlling in-context Machine Translation. 

\section{Limitations} 

This section details several limitations and ethical concerns associated with this work. 

\begin{itemize} 
\item While we have identified coherency of domain and document as a factor for in-context MT, we expect there should be other factors that could be more predictive of downstream performance, such as activation of attention patterns from source to target sentence during generation.  
 
\item We studied GPTNeo, Bloom and XGLM which have different training data but similar sizes. Due to GPU memory limitations we did not study larger models and it is not clear whether findings generalise to even larger models.

\item Although the TED talks dataset is a good overall testbed because it covers many topics and combines formal language and informal text, we did not quantify whether coherence is more likely to affect formal or informal language and might be studied with other datasets.

\item This paper focuses heavily on MT as a complex generative task to study coherence of context, and it is not immediately clear whether the findings would also generalise to other longer-context generation tasks such as document summarization or how this would affect simple classification.  
\end{itemize}

\section{Ethical Concerns}
Large LMs are known to hallucinate text content, potentially produce toxic speech or misinformation. While we did not observe this frequently in our experiments, we did not quantify the extent of this across various methods.  

\bibliography{main}
\bibliographystyle{acl_natbib}

\appendix

\include{Appendix}

\end{document}

%% file: tables/format_example.tex
\begin{table*}
\begin{small}
\begin{tabular}{lll}
\toprule
Translate English to French. & & \\
English: A discomfort which lasts .. & French: & Un malaise qui dure
 \\
English: HTML is a language for formatting & French: & HTML est un langage de formatage \\

... &  & ...\\
English: After you become comfortable with formatting .. & French: & \\
\bottomrule
\end{tabular}
\end{small} 
\caption{\small A single continuous input sequence presented to the model for decoding a single test source sentence ``After you become comfortable with formatting..''. Given the entire sequence as input, the model proceeds to generate the target sequence.}
\label{tab:format_example}
\end{table*}

%% file: tables/domain_results.tex
\begin{table*}[!t]

\resizebox{2\columnwidth}{!}{
\begin{tabular}{lcccc|cccc|cccc}
\toprule

& \multicolumn{4}{c}{GPTNeo2.7B} & \multicolumn{4}{c}{Bloom3B} & \multicolumn{4}{c}{XGLM2.9B} \\
Prompt / Test &  FLORES &     MED &    MTNT &     TED &  FLORES &     MED &    MTNT &     TED & FLORES &     MED &    MTNT &     TED \\

\midrule
FLORES & \textbf{24.6 }&  \textbf{19.7} &  \textbf{23.1} &  \textbf{24.6} & \textbf{36.7} &  28.5 &  28.5 &  31.1 & \textbf{ 29.3} &  20.9 &  24.7 &  \textbf{25.7} \\
MED & 23.0 &  19.2 &  21.1 &  23.2 & 34.5 &  \textbf{28.7} &  26.2 &  29.5 & 27.5 &  \textbf{21.4} &  22.9 &  24.4 \\
MTNT &  23.7 &  18.6 &  22.4 &  23.7 & 35.5 &  27.7 &  \textbf{29.1 }&  30.6 &  27.9 &  21.2 &  \textbf{25.0} &  25.4 \\
TED & 23.2 &  18.6 &  22.1 &  23.6 & 36.1 &  27.9 &  \textbf{29.1} &  \textbf{31.2} & 27.8 &  21.1 &  24.2 &  24.8 \\



\bottomrule
\end{tabular}}
\caption{Crosstable of BLEU scores from sampling and testing in different domains. We present the average BLEU scores across 5 randomly sampled prompt sets. The size of the prompt sets (number of translation pair examples) is 5. We bold the largest value column-wise.}
\label{tab:domain_results}
\end{table*}

%% file: tables/domain_length.tex
\begin{table}[h!]
\resizebox{\columnwidth}{!}{
\begin{tabular}{lllll}

Prompt / Test &      FLORES &         MED &        MTNT &         TED \\
\midrule

FLORES & - & - & - & - \\
MED   &  \rD 22.4\;(0.3) &  \rD 18.5\;(0.3) &  \rd 20.8\;(0.8) &  \rD 22.5\;(0.8) \\
MTNT  &  \rD 23.2\;(0.4) &  \rd 18.3\;(0.5) &  \rD 21.9\;(1.2) &  \rd 23.5\;(0.5) \\
TED   &  \rD 21.7\;(1.4) &  \rD 17.6\;(0.6) &  \rD 20.1\;(1.8) &  \rD 22.3\;(1.5) \\
\midrule
\multicolumn{5}{c}{5-10 words long sentences; GPTNeo 2.7B} \\
& & & & 
\end{tabular}}



\resizebox{\columnwidth}{!}{
\begin{tabular}{lllll}

Prompt / Test &  FLORES &     MED &    MTNT &     TED \\

\midrule
FLORES &  24.2\;(0.2)\rd  &  19.6\;(0.3) &   22.7\;(0.8)\rd &  24.3\;(0.5)\rd  \\
MED    &  22.9\;(0.6) &  19.3\;(0.1) &  21.1\;(0.9) &  22.8\;(0.7) \rd \\
MTNT   &   24.0\;(0.4) \gu &   18.9\;(0.6)\gu &  22.5\;(0.0) &   24.3\;(0.3)\gU \\
TED    &   23.8\;(0.4)\gU &   19.0\;(0.4)\gu &   22.9\;(0.2)\gU &  23.8\;(0.4) \\
\midrule
\multicolumn{5}{c}{15-20 words long sentences; GPTNeo 2.7B} \\
\end{tabular}}
\caption{Selecting for short source sentences (5-10 words) vs longer source sentences (15-20 words) as translation examples. \rd and \gu refers to differences $>0.3$, and \rD and \gU refers to differences $>0.5$ when compared  to the no-length filter scenario in \autoref{tab:domain_results}.}
\label{tab:domain_length}
\end{table}

%% file: tables/doc_level_results.tex
\begin{table*}[!t]
\centering
\resizebox{2\columnwidth}{!}{
\begin{tabular}{ll|lll|lll|lll||ll}
          &    &      \multicolumn{3}{c}{GPTNeo2.7B(BLEU)}  & \multicolumn{3}{c}{Bloom3B(BLEU)} & \multicolumn{3}{c}{XGLM2.9B(BLEU)} & \texttt{L2} & \texttt{coverage}   \\
   
          & In/outdoc & \texttt{en}$\rightarrow$\texttt{fr}         & en-pt & en-de & en-fr         & en-pt & en-de & en-fr            & en-pt & en-de  & - & -\\
          \midrule
\texttt{random}    & out &        26.3       &   27.1    &  16.6     &     35.2     &     35.5  &   7.9    &    27.1       &  26.7     &  18.9   & 1.35  & 0.31 \\
\texttt{nn}        & out &        26.8       &    26.9   &   16.9    &      35.1         &   35.1    &   8.2    & 25.6     &  26.6     &  18.3    & 0.98  &  0.49 \\
\texttt{bm25}      & out &        27.1       &   27.4   &   17.3    &       35.1        &   35.3    & \textbf{9.4}  & 25.3 &  27.0     &   18.4      & 1.21 &  0.75 \\
\texttt{bm25-s} & out &           27.2    &    27.5   &    17.4   &        34.8       &   34.9    &  9.1     &    25.6         &  27.4     &  18.7     & 1.25 & 0.80 \\
\texttt{random}    & within &    27.4      &   27.3    &   17.3    &        35.9       &  35.8     &   7.8    &  26.6     &  28.8     &  19.6    & 1.28 &  0.34 \\
\texttt{window}    & within &    \textbf{28.1} &  \textbf{28.3} & \textbf{17.9}  & \textbf{36.9}  & \textbf{37.0} &   8.8  &  \textbf{28.6}&  \textbf{31.6} &  \textbf{21.2}     & 1.22 & 0.40
\end{tabular}}
\caption{BLEU score comparison of similarity-based retrieval methods from out of document, and moving window (\texttt{window}) from within the document. \texttt{Coverage} (Rouge1-precision) refers to the word overlap between prompt source sentences and test source sentence. \texttt{L2} refers to the average L2 Euclidean distance between source prompt sentence embeddings and the test sentence embedding.}
\label{tab:doc_level_results}
\end{table*}


%% file: tables/shuffle_doc_level.tex
\begin{table}[t!]
\centering
\resizebox{0.8\columnwidth}{!}{
\begin{tabular}{lcccc}
\toprule
                retrieval &  bleu & \texttt{L2} & \texttt{Coverage} &  \texttt{ppl\_s} \\
\midrule

    \texttt{static} &  26.6 &  1.22   &  0.41  &    16.8 \\
 \texttt{random} &  27.4 & 1.28   & 0.31 &   14.9 \\
   \texttt{window} &  28.1 & 1.22    & 0.40 &  11.1 \\
   \texttt{shuffle}  &  28.3 & 1.22   &    0.40  &   12.0 \\
\bottomrule
\end{tabular}}
\caption{Ordering effects within document. All retrieval methods are within documnent.}
\label{tab:shuffle_doc_level}
\end{table}

%% file: tables/interference.tex
\begin{table}[!t]
\resizebox{\columnwidth}{!}{
\begin{tabular}{l|ccccc}
 & \texttt{random} & \texttt{bm25} & \texttt{bm25-s} & \texttt{nn} & \texttt{window} \\
 \midrule
Positive & 0.56 &      \textbf{0.62} &  0.61 &     0.6 & \textbf{0.62} \\
Negative & 0.32 &      0.31 &  0.31 &    0.32 & \textbf{0.29} \\
No Change & 0.12 &      0.07 &  0.08 &    0.08 &       0.09 \\ 
\end{tabular}}
\caption{Positive, Negative and No change (proportions) in BLEU scores across different prompt selection methods. For positive row, higher is better. For negative row, lower is better.}
\label{tab:interference}
\end{table}

%% file: Appendix.tex
\section{Resources}
\paragraph{Software}
\begin{itemize}
    \item Implementation of Nearest Neighbor Retrieval with FAISS library \cite{johnson2019billion}
    \item Huggingface library was used for LLMs, model weights and calculation of perplexity.
\end{itemize}
\paragraph{Hardware}
\begin{itemize}
\item All experiments can be run with a single NVIDIA-TITAN RTX GPU (24GB).
\end{itemize}
\paragraph{Datasets} All datasets for the experiments are open-source.

\section{Local Coherence (nprompts=15)}
\label{appendix:exp_coherence}
Ablation experiments for \autoref{sec:exp_coherence} prompt set size of 15 shown in \autoref{tab:doc_level_results_nprompts15}. The same trend is observed for 5 and 15 prompts.

\input{tables/doc_level_results_nprompts15.tex}

\section{Domain vs Perplexity}
\label{appendix:domain_ppl}
We report the perplexity of the source sentences when randomly sampling and testing from different domains.
Although there is no widely accepted measure of \textit{general coherence}, we can formulate this with respect to the particular model being studied. We consider the model's conditional perplexity of the test sentence given the context. Perplexity is a widely used measure of suprisal in text and has also been used as a measure in topic coherence (reference).  Concurrent work by \citet{gonen2022demystifying} argue that total perplexity of the input sequence is related to In-context performance. We report the conditional perplexity from sampling and testing in different domains for GPTNeo2.7B in \autoref{tab:ppl_domain_gptn} and Bloom3B in \autoref{tab:ppl_domain_bloom}. We did not report XGLM2.9B because the model log likelihood is very poorly calibrated.

\begin{table}[h!]
    \centering
\begin{tabular}{lllll}
\toprule
Prompt / Test  & FLORES &   MED &  MTNT &   TED \\
\midrule
FLORES &   \textbf{21.3} &  24.5 &  54.9 &  25.5 \\
MED    &   24.1 &  \textbf{16.2} &  62.4 &  27.0 \\
MTNT   &   25.4 &  26.9 &  \textbf{40.8} &  23.3 \\
TED    &   24.0 &  24.9 &  52.2 &  \textbf{19.6} \\
\bottomrule
\end{tabular}
\caption{Source sentence perplexity conditioned on prompts randomly sampled from the domain computed with GPTNeo2.7B. Lower perplexity indicates greater coherence.}
\label{tab:ppl_domain_gptn}
\end{table}

\begin{table}[h!]
    \centering
\begin{tabular}{lllll}
\toprule
Prompt / Test  & FLORES &   MED &  MTNT &   TED \\
\midrule
FLORES &   \textbf{21.5} &  23.4 &  61.8 &  28.5 \\
MED    &   25.1 &  \textbf{16.3} &  74.8 &  33.4 \\
MTNT   &   25.2 &  25.7 &  \textbf{47.2} &  26.1 \\
TED    &   24.1 &  23.5 &  60.1 &  \textbf{22.1} \\
\bottomrule
\end{tabular}
\caption{Source sentence perplexity conditioned on prompts randomly sampled from the domain computed with Bloom3B. Lower perplexity indicates greater coherence.}
\label{tab:ppl_domain_bloom}
\end{table}

\section{Local Coherence Shuffle Effects}
\label{appendix:shuffle}
BLEU scores for comparing \texttt{window} with other baselines, accompanying appendix section to \autoref{sec:exp_shuffle} which reports BLEU scores for GPTNeo2.7B \texttt{en}$\rightarrow$\texttt{fr}. we find that results generalise across several models and languages that we further investigated. \input{tables/shuffle_doc_level_appendix.tex}

%% file: tables/doc_level_results_nprompts15.tex
\begin{table*}[!t]
\centering
\resizebox{2\columnwidth}{!}{
\begin{tabular}{ll|lll|lll|lll|}
          &    &      \multicolumn{3}{c}{GPTNeo2.7B(BLEU)}  & \multicolumn{3}{c}{Bloom3B(BLEU)} & \multicolumn{3}{c}{XGLM2.9B(BLEU)}   \\
   
          & In/outdoc & en-fr         & en-pt & en-de & en-fr         & en-pt & en-de & en-fr            & en-pt & en-de  \\
          \midrule
\texttt{random}    & out &        27.2       &   27.3    &  16.9     &     35.3     &     35.4  &   8.0    &    29.2       &  31.2     &  20.7    \\
\texttt{nn}        & out &        27.3       &    28.2   &   17.1    &      35.6         &   35.9    &   9.1    & 30.0     &  32.0     &  21.6   \\
\texttt{bm25}      & out &        27.9       &   29.0   &   17.4    &       36.1        &   36.4    & \textbf{10.8}  & 31.2 &  33.0     &   22.2    \\
\texttt{bm25-s} & out &           27.7    &    29.1   &    17.3   &        35.2       &   36.0    &  9.1     &    29.8         &  32.0     &  21.6  \\
\texttt{random}    & within &    28.1      &   29.2    &   17.6    &        36.8       &  37.3     &   8.9    &  30.9     &  33.3     &  22.3    \\
\texttt{window}    & within &    \textbf{28.9} &  \textbf{29.8} & \textbf{18.2}  & \textbf{37.8}  & \textbf{38.1} &   9.6  &  \textbf{31.7}&  \textbf{34.4} &  \textbf{23.0} 
\end{tabular}}
\caption{BLEU score comparison of similarity-based retrieval methods from out of document, and moving window (\texttt{window}) from within the document. 15 prompt examples used.}
\label{tab:doc_level_results_nprompts15}
\end{table*}

%% file: tables/shuffle_doc_level_appendix.tex
\begin{table}[t!]
\centering
\resizebox{\columnwidth}{!}{
\begin{tabular}{p{0.15\columnwidth}p{0.20\columnwidth}p{0.20\columnwidth}p{0.15\columnwidth}p{0.2\columnwidth}}
\toprule
                 &  GPTNeo (en-pt) & GPTNeo (en-de) & XGLM (en-fr) & Bloom (en-fr)  \\
\midrule

\texttt{static} &  27.1 & 16.8 & 27.7 & 34.9 \\
 \texttt{random} &  27.3 & 17.3 & 26.6 & 35.9 \\
   \texttt{window} &  28.3 & 17.9 & 28.5 & 36.9 \\
   \texttt{shuffle}  &  28.5 & 17.9 & 28.7 & 36.9 \\
\bottomrule
\end{tabular}}
\caption{BLEU scores for different ordering effects within document. All retrieval methods are within document. }
\label{tab:shuffle_doc_level_appendix}
\end{table}